\documentclass[runningheads]{llncs}
\usepackage[T1]{fontenc}
\usepackage{graphicx}
\DeclareUnicodeCharacter{2212}{-}
\usepackage{enumerate}

\begin{document}
\title{Representation Learning on Graphs to Identifying Circular Trading in Goods and Services Tax}
\titlerunning{Identifying Circular Trading in Goods and Services Tax}
\author{
Priya Mehta\inst{2}
\and
Sanat Bhargava\inst{1}\orcidID{0000-0002-2169-887X} \and
M. Ravi Kumar \inst{1} \and
K. Sandeep Kumar\inst{1} \and
Ch. Sobhan Babu\inst{1} 
}
\authorrunning{Priya Mehta et al.}
\institute{Department of Computer Science and Engineering, Indian Institute of Technology, Hyderabad, India \\
\email{\{sanat.bhargava,CS18MTECH11028, sobhan,sandeep\}@cse.iith.ac.in}
\and
Welingkar Institute of Management Development and Research, Mumbai, India \\
\email{priya.mehta@welingkar.org}}
\maketitle              % typeset the header of the contribution
\begin{abstract}
Circular trading is a form of tax evasion in {\it Goods and Services Tax} where a group of fraudulent taxpayers (traders) aims to mask illegal transactions by superimposing several fictitious transactions (where no value is added to the goods or service)  among themselves in a short period. Due to the vast database of taxpayers, it is infeasible for authorities to manually identify groups of circular traders and the illegitimate transactions they are involved in. This work uses big data analytics and graph representation learning techniques to propose a framework to identify communities of circular traders and isolate the illegitimate transactions in the respective communities. Our approach is tested on real-life data provided by the Department of Commercial Taxes, Government of Telangana, India, where we uncovered several communities of circular traders.

\keywords{Graph Representation Learning \and Graph Clustering \and Fraud Detection \and Circular Trading \and Goods and Services Tax \and Value Added Tax}
\end{abstract}
\section{Introduction}
\begin{figure}
\begin{center}
\includegraphics[width=.8\textwidth]{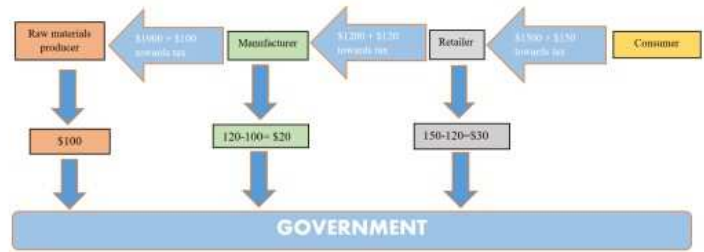}
\caption{The GST system} \label{fig1}
\end{center}
\end{figure}

Taxes can be broadly classified into two categories, based on how they are collected, namely, direct taxes and indirect taxes.  This work proposes an approach to identify groups of taxpayers (traders) that engage in evasive activity called {\it circular trading} in the indirect taxation system.   In indirect taxation systems, the tax is imposed successively at every point of the value chain, and the amount of tax levied is contingent on the value added to goods at the corresponding point in the value chain. Tax is levied at each point of the value chain, such that tax paid on the requisite goods and services (input tax) for that point offsets the tax levied on the sales (output tax). Figure \ref{fig1} shows how the tax is collected incrementally in this system. 

In this example, the manufacturer purchases some raw material of value \$1000 from the raw material dealer by paying \$100 as tax at a rate of 10\%. The raw materials dealer remits this tax to the government. The retailer purchases the processed goods from the manufacturer for, say, \$1200 and pays \$120 to the manufacturer as tax at the same rate of taxation. The manufacturer pays the government the difference between the tax he had collected from the retailer and the tax he has paid to the raw materials producer, i.e., \$120 - \$100 = \$20. The consumer then buys the finished goods from the retailer for \$1500 by paying a tax of \$150. Following the same argument, as given in the previous steps, the retailer pays the government \$30(i.e.,\$150 − \$120).

It is easy to see that the total tax received by the government is \$150, and the entirety of the tax burden is indirectly shifted onto the consumer of goods. Hence, the raw materials dealer, the manufacturer, and the retailers collect the tax on behalf of the government. 

\subsection{Circular Trading in GST and VAT}
\begin{figure}
\begin{center}
\includegraphics[width=.6\textwidth]{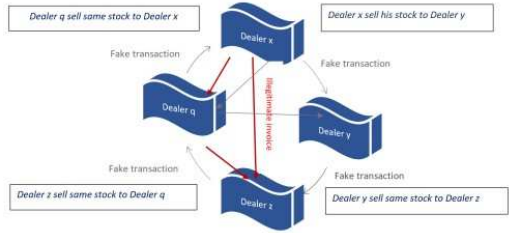}
\caption{Circular Trading amongst dealers} \label{fig2}
\end{center}
\end{figure}

In most of the detected tax evasion cases, business dealers (taxpayers) defrauded the authorities by deliberately manipulating the actual business transactions in their tax returns to maximize the amount of profit gained by evading tax. Invoice trading is a tax evasion method, in which a dealer sells their goods to the end-user and collects the corresponding tax  without issuing an invoice of sale. Later, they issue a illegitimate invoice to a third party, implying that the end-user's tax is attributable to the third party. This enables the third party to increase their input tax credit, therefore minimizing the tax they have to pay in the form of cash (the difference between the tax they collected at the time of sales and the tax they paid at the time of purchases) to the government. To mask these illegitimate transactions, malicious dealers devised an elaborate scam, where many shell companies are created to doctor the title of goods in the first place and subsequently make fictitious (fake) transactions among themselves to bypass the detection system. Tax evaders show high-valued fake sales and purchases among themselves and dummy dealers (shell firms) without adding any significant value to the product itself, as depicted in Figure \ref{fig2}. In Figure \ref{fig2}, edges from x to q, x to z, and q to z depict illegitimate transactions. Dealers z practice invoice trading to minimize his respective tax liabilities. To befuddle the tax enforcement authorities, these dealers superimpose several fake transactions (depicted using grey lines in fig \ref{fig2}) on these illegitimate transactions (depicted using red lines in fig \ref{fig2}). A key characteristic of superimposed fake transactions is that the tax liability resulting from these transactions is zero, i.e., the amount of tax paid on these purchases is equal to the amount of tax collected on these sales. Since there is no value-addition due to the these transactions, the dealers (taxpayers) do not pay any tax on these fictitious transactions.
Furthermore, this behavior tends to confuse the authorities due to the large volume of fake transactions. The number of fake transactions is much higher than that of genuine transactions. This method of tax evasion is known as circular trading. 

As a consequence of the vast size and ever-expanding nature of the tax department’s database of taxpayers, it is impractical for the tax authorities to identify the groups of evaders that practice circular trading. The problem is further compounded by the fact that elaborate sequences of sales and purchase transactions in circular trading and the unknown identity of the taxpayers doing these manipulations make it difficult to identify and isolate groups that indulge in circular trading. These challenges call for sophisticated big data and graph machine learning techniques. We implemented these algorithms for the Commercial Taxes Department, Government of Telangana, India.  

\section{Related Work}
Most of the prior work focuses on detecting circular trading in the stock market. In \cite{palshikar2008collusion}, a graph clustering algorithm is presented for detecting collusion sets in the stock market using Dempster–Schafer's theory of evidence. In \cite{wang2012detecting}, authors proposed a method to detect potential collusive cliques by computing the coefficient of correlation between two suitable unified aggregated time series of signed order volume and then combining the connected components from multiple sparsified weighted graphs constructed by using correlation matrices where each correlation coefficient is higher than a threshold specified by the user. In \cite{islam2009approach},  authors proposed an approach to detect collusion sets using Markov Clustering Algorithm. This method can detect purely circular collusion as well as cross-trading collusion. In \cite{mehta2017graph}, authors proposed the {\it Bottleneck edge computation algorithm}, which is applied to remove cycles from circular trading communities provided that the communities are known beforehand.  Machine learning tasks over nodes and edges in graph require careful effort in extracting features used by machine algorithms. In machine learning, feature learning or representation learning is a set of techniques that allows a system to automatically discover the representations needed for feature detection or classification from raw data.  In \cite{grover2016node2vec}, authors proposed {\it node2vec} algorithm, which learn a mapping of nodes to a low-dimensional space. In \cite{HamiltonYL17}, authors provided a conceptual review of key advancements in  representation learning on graphs, including matrix factorization-based methods, random-walk based algorithms, and graph convolutional networks.

\section{Description of Data Set and Graph Construction Procedure}
\label{dataset}
\begin{table}
\caption{GSTR-1 DATABASE}\label{tab1}

\begin{tabular}{|l|l|l|l|l|}
\hline
S.No &  Seller & Buyer & Time of Transaction & Sales (in Rupees)\\
\hline
1 &  Dealer A & Dealer B & 2021/03/03/10:10 & 14000\\
2 &  Dealer C & Dealer D & 2021/03/03/10:40 & 17000\\
3 &  Dealer A & Dealer D & 2021/03/10/09:30 & 12000\\
4 &  Dealer B & Dealer C & 2021/03/11/00:30 & 15000\\
\hline
\end{tabular}
\end{table}

The Goods and Services Tax Return 1 (GSTR-1) is a document that each registered tax payer needs to file every month/quarter. It must contain the details of all sales and supply of goods and services made by the tax payer during the tax period. We have taken GSTR-1 invoices data from March 2021 to December 2022. Number of invoices are 77837538, number of taxpayers are 388448 and  size of the data set is 7GB.

 Table ~\ref{tab1} depicts a few fields of the sales transactions data set (GSTR-1) used. Construct an edge labelled directed multigraph called {\it sales flow graph} using this data set. Each vertex in this graph corresponds to an individual dealer. Each directed edge denotes one sales transaction in the data set.  Each edge contains two attributes, where the first one is the time of sales of the corresponding transaction and the second one is the monetary outflow associated with this transaction.  Due to the complexity and the size of this graph, community detection is a non-trivial task.

\section{Identifying Communities in Sales Flow Graph}
\subsection{Node2vec}
Node2vec is a semi-supervised algorithm  which  assigns an embedding (essentially a vector in n-dimensional space) for every node in the graph, such that the relative position of the node is preserved. These embeddings allow us to quantify the similarity or the dissimilarity between nodes, which is crucial for clustering nodes of the graph. It is worth noting that by tuning the hyperparameters appropriately we can get node embeddings such that the nodes which belong to the same neighbourhood, have similar embeddings. The quality of embeddings is influenced by the choice of following hyperparameters.

\begin{itemize}

\item \textbf{Return parameter(p)}: Controls the probability of revisiting a node. We set p=1 to ensure moderate exploration and steer clear of two-hop redundancy.

\item \textbf{In-out parameter(q)}: Controls the bias towards "inward" and "outward" nodes. We set q=1/2 to increase the likelihood of visiting nodes further away from the current node. Our choice of p and q ensures a Depth-First Sampling of the neighbourhood.

\item \textbf{Length of RandomWalk}: The number of nodes to traverse from the current node to sample probabilities of every node in the neighbourhood of the current node.

\item \textbf{Number of RandomWalks}: Number of Random Walks to carry out for each node.

\item \textbf{Dimensions}: Number of dimensions of the vector space containing the embedding vectors. Conventionally, this value is taken to be $\sqrt(number ~of~nodes)$.
    
\end{itemize}

\subsection{Clustering using DBSCAN}
Once we have obtained the embeddings for every node, we find densely connected nodes within the graph. Mathematically, this reduces to clustering embedding vectors that are close to each other in n-dimensional space. For this, we use  DBSCAN algorithm to find dense regions within the vector space using {\it cosine similarity measure}. As with node2vec, the effectiveness of clustering depends on the choice of hyperparameters.  Two important hyper parameters required for DBSCAN are {\it epsilon} (“eps”) and minimum points (“MinPts”). The parameter {\it eps} defines the radius of neighborhood around a point. The parameter {\it MinPts} is the minimum number of neighbors within “eps” radius. 

\subsection{Algorithm to find communities}

\begin{enumerate}[\bfseries Step 1:]
    \item Create an edge labelled  directed multi graph from the sales transactions data set as explained in Section \ref{dataset}.
    \item Convert the above graph to an edge-weighted undirected graph as explained in \cite{MehtaMKSBR19}.
    \item Generate embeddings for each node of the undirected graph using node2vec algorithm.
    \item Apply DBSCAN algorithm to find clusters of nodes that are densely connected together.
\end{enumerate}

\section{Results and Conclusion}
\begin{figure}[!tbp]
  \centering
  \begin{minipage}[b]{0.4\textwidth}
    \includegraphics[width=\textwidth]{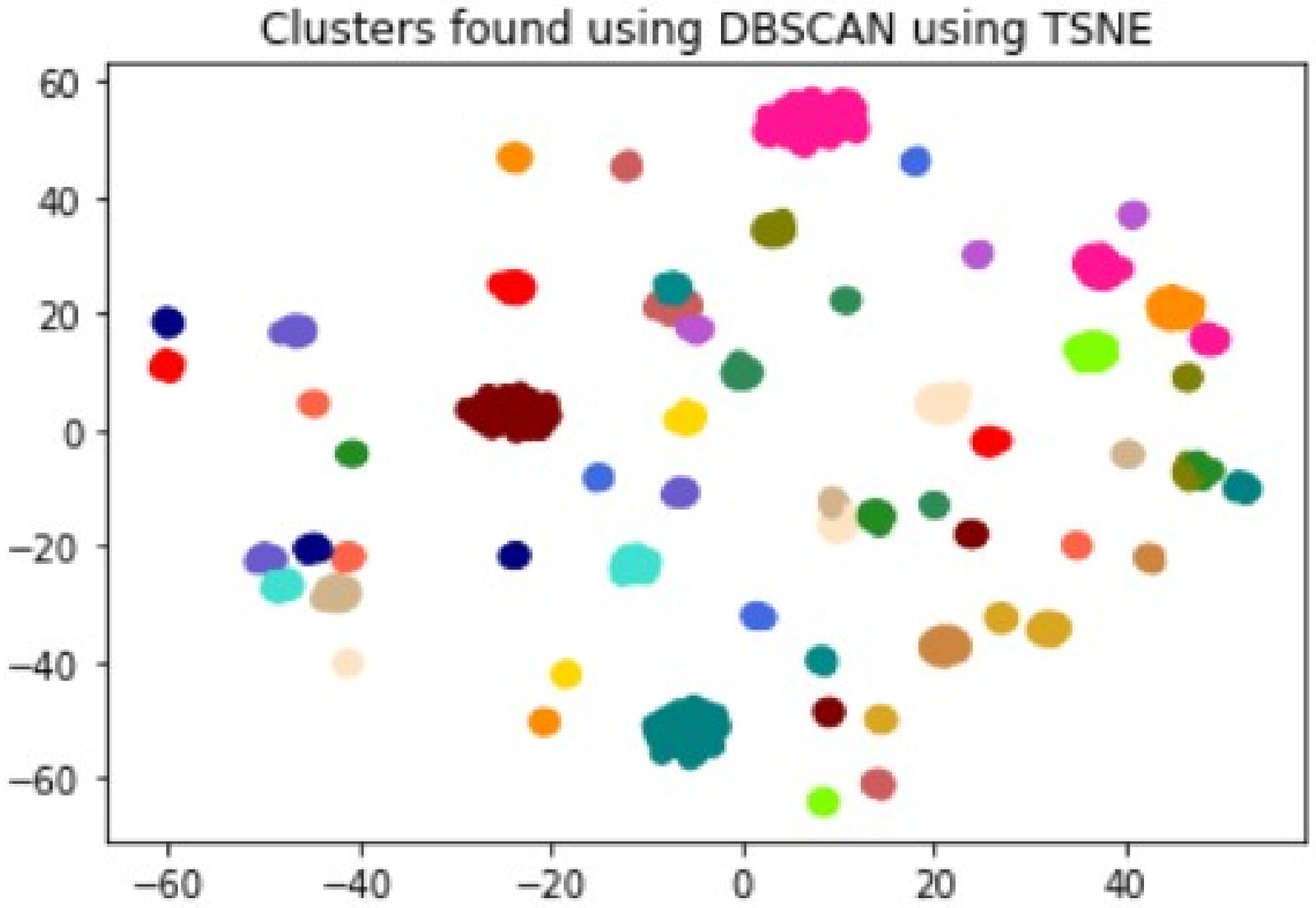}
    \caption{Clusters obtained from DBSCAN}
    \label{fig3}
  \end{minipage}
    \hfill
  \begin{minipage}[b]{0.4\textwidth}
    \includegraphics[width=\textwidth]{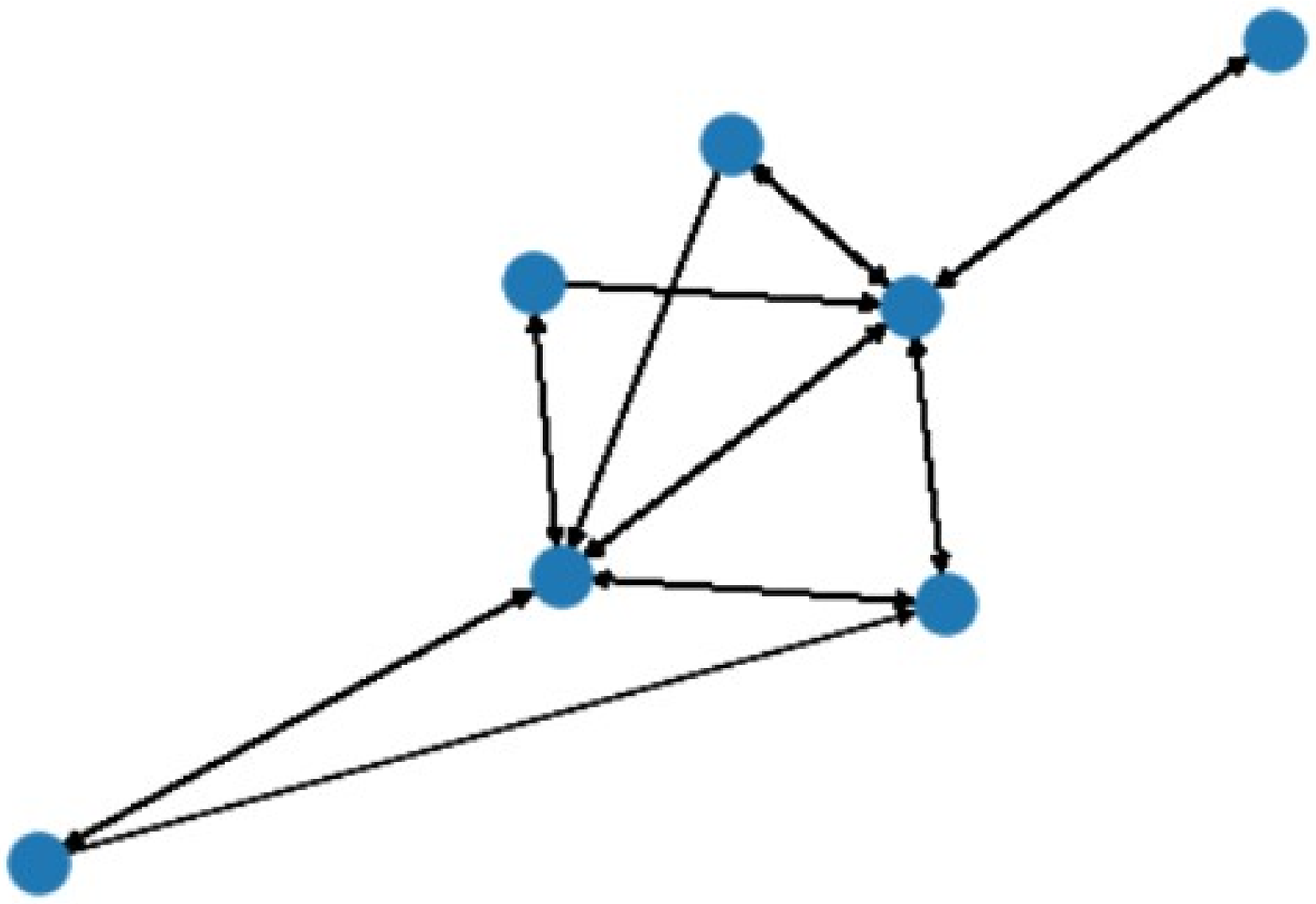}
    \caption{A community within the sales flow graph}
    \label{fig4}
  \end{minipage}
\end{figure}

The  Figure \ref{fig3} shows the clusters obtained after applying the DBSCAN algorithm.  The  Figure \ref{fig4} shows one cluster in detail. Here, we present a state-of-the-art methodology to obtain communities from a large and convoluted graph. Once we have obtained the communities, we can use the algorithms described in \cite{mehta2017graph} to isolate fraudulent transactions within the community.

%\begin{figure}
%\includegraphics[width=.3\textwidth]{dbscan.eps}
%\caption{Clusters obtained from DBSCAN} \label{fig3}
%\end{figure}
%\hfill
%\begin{figure}
%\includegraphics[width=.3\textwidth]{commdet.eps}
%\caption{A community within the sales flow graph.} \label{fig4}
%\end{figure}

\noindent { \bf Acknowledgements: }  We would like to express our gratitude towards the government of Telangana, India, for allowing us access to the Commercial Taxes Data set for this work. 

\bibliographystyle{splncs04}
\bibliography{citations}
\end{document}